\documentclass[10pt,twocolumn,letterpaper]{article}

\usepackage{cvpr}
\usepackage{times}
\usepackage{epsfig}
\usepackage{graphicx}
\usepackage{amsmath}
\usepackage{amssymb}

\usepackage[pagebackref=true,breaklinks=true,letterpaper=true,colorlinks,bookmarks=false]{hyperref}

\cvprfinalcopy % *** Uncomment this line for the final submission

\def\cvprPaperID{****} % *** Enter the CVPR Paper ID here

\ifcvprfinal\fi
\begin{document}

%%%%%%%%% TITLE
\title{Finding your Lookalike: \\Measuring Face Similarity Rather than Face Identity}

\author{Amir Sadovnik, Wassim Gharbi, Thanh Vu\\
Lafayette College\\
Easton, PA\\
{\tt\small \{sadovnia,gharbiw,vut\}@lafayette.edu}
% For a paper whose authors are all at the same institution,
% omit the following lines up until the closing ``}''.
% Additional authors and addresses can be added with ``\and'',
% just like the second author.
% To save space, use either the email address or home page, not both
\and
Andrew Gallagher\\
Google Research\\
Mountain View, CA\\
{\tt\small agallagher@google.com}
}

\maketitle
%\thispagestyle{empty}

%%%%%%%%% ABSTRACT
\begin{abstract}

        Face images are one of the main areas of focus for computer vision, receiving on a wide variety of tasks. Although face recognition is probably the most widely researched, many other tasks such as kinship detection, facial expression classification and facial aging have been examined. In this work we propose the new, subjective task of quantifying perceived face similarity between a pair of faces. That is, we predict the perceived similarity between facial images, given that they are \textit{not} of the same person. Although this task is clearly correlated with face recognition, it is different and therefore justifies a separate investigation. Humans often remark that two persons look alike, even in cases where the persons are not actually confused with one another. In addition, because face similarity is different than traditional image similarity, there are challenges in data collection and labeling, and dealing with diverging subjective opinions between human labelers. We present evidence that finding facial look-alikes and recognizing faces are two distinct tasks. We propose a new dataset for facial similarity and introduce the Lookalike network, directed towards similar face classification, which outperforms the ad hoc usage of a face recognition network directed at the same task.

\end{abstract}

%%%%%%%%% BODY TEXT
\section{Introduction}

\label{sec:intro}

Have you ever seen an actor or actress and thought that they looked similar to someone that you know? Although you would clearly not confuse the two as being the same person, there might be some characteristics which may remind you of a certain person. You might be able to describe the attributes of why these people look alike, or the similarity may not even be nameable. Although your internal ``face recognition algorithm'' understands that these two individuals are different, something is still marking them as similar.  

\begin{figure}
        
        \centerline{\includegraphics[width=8.5cm]{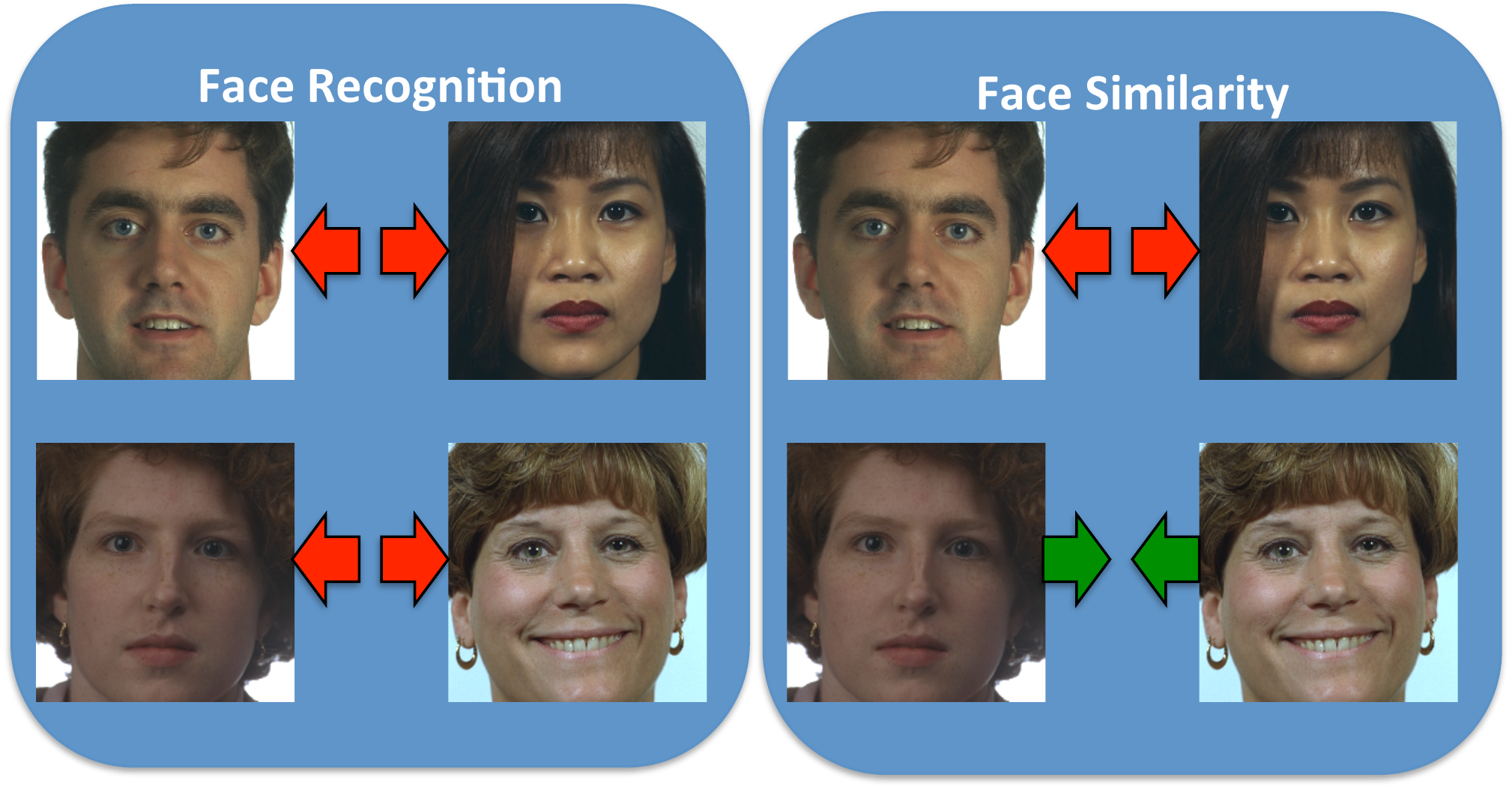}}
        
        %  \vspace{2.0cm}
        
        \caption{Most face recognition algorithms embed faces in a space such that different images of the same person are close, while faces of different people are far. This goal is agnostic to the perceived similarity of the faces. In this work, we learn a Lookalike network that maps faces to an embedding space where the distance between face images of different people depends upon perceived similarity to human observers.}
        
        \label{fig:firstpage}
        
\end{figure}

Is this notion of face similarity captured as a natural side effect by training a face recognizer that performs fine-grained instance recognition? Recent face recognition algorithms are trained with identity as classes, and nothing explicitly captures the idea that some people look more like others. As shown in Fig. \ref{fig:firstpage} most of these algorithms are trained by learning an embedding space in which images of the same person are encouraged to be close, while images of different people are far, without regard for how similar they appear. That is, so long as images of different people are ``far enough'', there is no motivation or reward for arranging faces in the embedding space according to the perceived similarity of the faces. 

Besides the interesting question of how face recognition and face similarity are related, there are also obvious applications for which a lookalike network would be better suited. For example, applications such as Microsoft's CelebsLike.Me website or Google's Arts \& Culture App have become popular by presenting a user with people similar to themselves. In addition, this type of network can be well-suited for movie casting to choose actors who appear similar to real-life people. 

Using this intuition we present what we believe is the first paper to investigate face similarity as an independent task. First, we show that this intuition is correct by collecting a dataset which examines the relationship between face similarity and face recognition. We then present the lookalike network which learns an embedding space specifically for this task. Finally, we present results which show that our lookalike network successfully learns how to measure face similarity, and is able to generalize to other datasets as well.

%Face recognition is one of the most well researched problems in computer vision. The problem of being able to recognize a person by his face given different poses, expressions, lighting conditions and facial accessories has been tackled using a variety of methods such as eigen-faces, clustering, and more recently,  convolution neural networks.

\section{Previous Work}

\label{sec:previous}

Although we propose the new task of learning facial similarity, our work is highly related to a few different subjects that are well-studied. First, since we are dealing with facial images, our work is highly related to the face recognition task. However, as we move from identity-as-category into a ranking approach, our work is also closely related to image similarity and metric learning. In this section we describe the prior work in these fields.

\subsection{Face Recognition}

Face recognition is one of the most widely researched topics in computer vision. One of the first papers to tackle the subject appeared in 1966 \cite{bledsoe1966model}, although that work required manually derived facial measurements (such as the locations of the corners of the eyes, the top of the nose, etc.) This early work already shed light on some of the later challenges of face recognition such as pose and expression invariance.

As the field progressed, features began to be extracted directly from image pixels. Some work  used dimensional reduction techniques on the pixels themselves for classification. Some examples include Eigenfaces \cite{turk1991face}, Fisherfaces \cite{belhumeur1997eigenfaces} and Laplacian Faces \cite{he2005face}. Other works described more advanced features based on ones that worked for images in other domains. Examples include the use of SIFT \cite{ke2004pca}, histograms of oriented gradients \cite{deniz2011face} and  Local Binary Patterns \cite{ahonen2006face}.

Recent years have seen the use of deep neural networks into almost all subproblems of computer vision. One of the main advantages in using these networks is that there is no longer a need to design features by hand, allowing the network to learn appropriate features for a specific task. Convolutional networks have also achieved state of the art results in face recognition. Taigman \emph{et al.} \cite{taigman2014deepface} propose a deep neural network for face verification. Their network requires the images to be aligned to facial landmarks before applying the network. Schroff \emph{et al.} \cite{schroff15} trained on unaligned images using a large database and triplet loss. Parkhi \emph{et al.} \cite{Parkhi15} managed to achieve similar results while training a deeper network and a much smaller training set.

Although there has been extensive work on face recognition as far as we know there has been no work that specifically addresses the concept of face similarity. However, there have been a few commercial applications which allow you to upload an image and search a celebrity database for the most similar images. For example, Mircosoft Inc. has a website called www.celebslike.me, which provides this service. Although they do not describe a specific algorithm they use for this task, they do link to the MSR-Celeb-1M paper \cite{guo2016ms}. This paper also addresses the task of face recognition, and therefore it can be assumed that similarity is measured by a network trained for face identity verification.

\subsection{Similarity Learning}

Measuring the similarity between two images has been an important task in image retrieval and  computer vision fields. Traditionally, many works used the distance between low-level feature representations to infer similarity. Features such as texture \cite{varma2005statistical,ma1996texture}, shape \cite{belongie2002shape}, or other features such as SIFT \cite{lowe2004distinctive} can be extracted from an image and compared to each other. Although this can yield good results for simple images, it tends to not perform as well for general images since these low-level features do not capture the high-level semantic concepts necessary to measure similarity as perceived by a human.

To capture these higher-level concepts, other papers tried to represent images using classifier outputs \cite{wang2009learning,rasiwasia2007bridging}. The main idea is that since these classifiers extract higher level information, they can be used for the image similarity task on a wider range of images. Although these do yield better results, the use of strictly semantic information eliminates the concept of visual similarity. For example, \cite{deselaers2011visual} examine the relationship between semantic and visual similarity and show that both are important.

Training on triplets of images to learn metric distances appears to have been introduced in \cite{Frome2006,Frome2007}. A triplet of images is comprised of two from the same category that are considered more similar, and one from a different category that is considered less similar to the others for learning a max-margin distance function over triplets. With the recent success of deep neural networks on many different tasks, they have been used for image similarity as well. The general idea in these works is to learn an embedding space in which distances between images correspond to similarity.  For example, \cite{wang2014learning} proposes training a deep neural network using a triplet loss to learn this embedding space, which in turn motivated the FaceNet triplet loss \cite{schroff15}.

Our work is similar to these more recent deep learning approaches with a few key extensions. First, we focus on face similarity rather than general image similarity. Rather than an instance classification problem, ours is clearly one of determining similarity. This is an important distinction since humans have dedicated neural processing for faces \cite{nineteenthings}   differently that other images; therefore, we believe facial similarity should be treated separately from general image similarity as well. In addition, datasets that contain subject identity are not sufficient for training a facial look-alike network. We collect our own ground truth specific for this task by mining human opinions of who looks like whom. The collection of this dataset adds additional complexities and opportunities. Finally, since face similarity is somewhat subjective, we integrate the disagreement between workers into our learning.

\section{Measuring The Difference Between Similarity and Recognition}

\label{sec:difference}

As described in Sec. \ref{sec:intro}, facial similarity and face recognition are related, yet distinct tasks. Although we should expect them to be highly correlated, our initial hypothesis is that a network trained for one specific task would not yield optimal results for the other. Therefore, in order to verify this hypothesis, we conducted an experiment to show that this is true, and show where the two tasks differ. More specifically, we use results from a state of the art face recognition algorithm, and show why it does not yield optimal results for a facial similarity task.

\subsection{Face Recognition Algorithm}

\label{sec:recognition}

To test our hypothesis we used the VGG-Face CNN descriptor \cite{Parkhi15}. This is descriptor is extracted by using a ``very deep'' convolutional neural network made out of 11 layers, the first 8 of which are convolutional and the last 3 of which are fully connected. The advantage of using this architecture appears to be that the network can achieve state of the art results with relatively small datasets. For example, in \cite{Parkhi15} they show comparable results to Facenet \cite{schroff15}  while using less than 1\% of the training samples.

In \cite{Parkhi15}, two different ways of training the network are discussed. One method is to learn the embedding using triplet loss. The other is to train it as a classification network, using a softmax layer during training with a loss related to the identity of the face. After the network is trained, the final classification layer is removed and the penultimate layer is used as the embedding for face recognition. Since similar results were achieved in each case, we use the second method. Using a pre-trained network we verify its accuracy by testing on LFW \cite{LFWTech} using their standard training and testing split. We achieve an ROC-AUC of 0.9773, on par with most recent face recognition results .

\subsection{Data Collection}

\label{subsec:data}

To test the performance of this facial recognition network at the task of facial similarity, we need to collect a new dataset that captures human opinions on who looks like whom in a face dataset. Using Amazon Mechanical Turk, we ask workers to compare two pairs of faces and choose the pair that looks more alike.

For this task we decide to use the Color-Feret Dataset \cite{phillips98,phillips00}.  Although the Color-Feret dataset is considered an easy dataset for face recognition (compared to more recent ones such as LFW) we decide to use it for this task for a few reasons. First, since we trying to examine how well a descriptor extracted from a face recognition algorithm performs on face similarity, we wish to have a dataset with excellent face recognition performance. That is, we want to ensure that the reason our face recognition descriptor does not perform well on similarity is not because it is not doing a good job on face recognition in the first place, but because the two tasks are inherently different. Second, since the photos are taken in controlled settings we believe it would be easier for a labeler to make a judgment on how similar the faces appear to each other.  Finally the dataset is well organized, and identities are guaranteed to be unique and accurate.

Since we wish to use this dataset to examine how well the VGG Face Descriptor performs on this task, we prefer to intentionally select specific images for comparison which will allow us to learn the most about the relationship between face recognition and face similarity. We select a single image per identity from the dataset (for simplicity we select the simple forward facing image with neutral expression) because we are interested in face similarity and not recognition. Then, we find the distance between the VGG face descriptors of all images in the dataset.

We then bin all pairs into 10 different bins based on their distance in the embedded space. For example all pairs of images whose Euclidean distance between them is 1.2-1.25 are in one bin, pairs whose distance is 1.25-1.3 would be in another bin, etc. We wish to compare pairs from a certain bin to pairs in all other bins. For example, if the face recognition distance is a good proxy for facial similarity, we would expect pairs which are in bin 1.1-1.15 to appear more similar than pairs in the 1.2-1.25 bin. We therefore select 100 test cases from each bin for labeling. Since we have 10 bins, and we are comparing each bin to all others, we have a total of $100 \times {{10}\choose{2} }= 4500$ pairs of pairs.

\begin{figure}

        \centerline{\includegraphics[width=8.5cm]{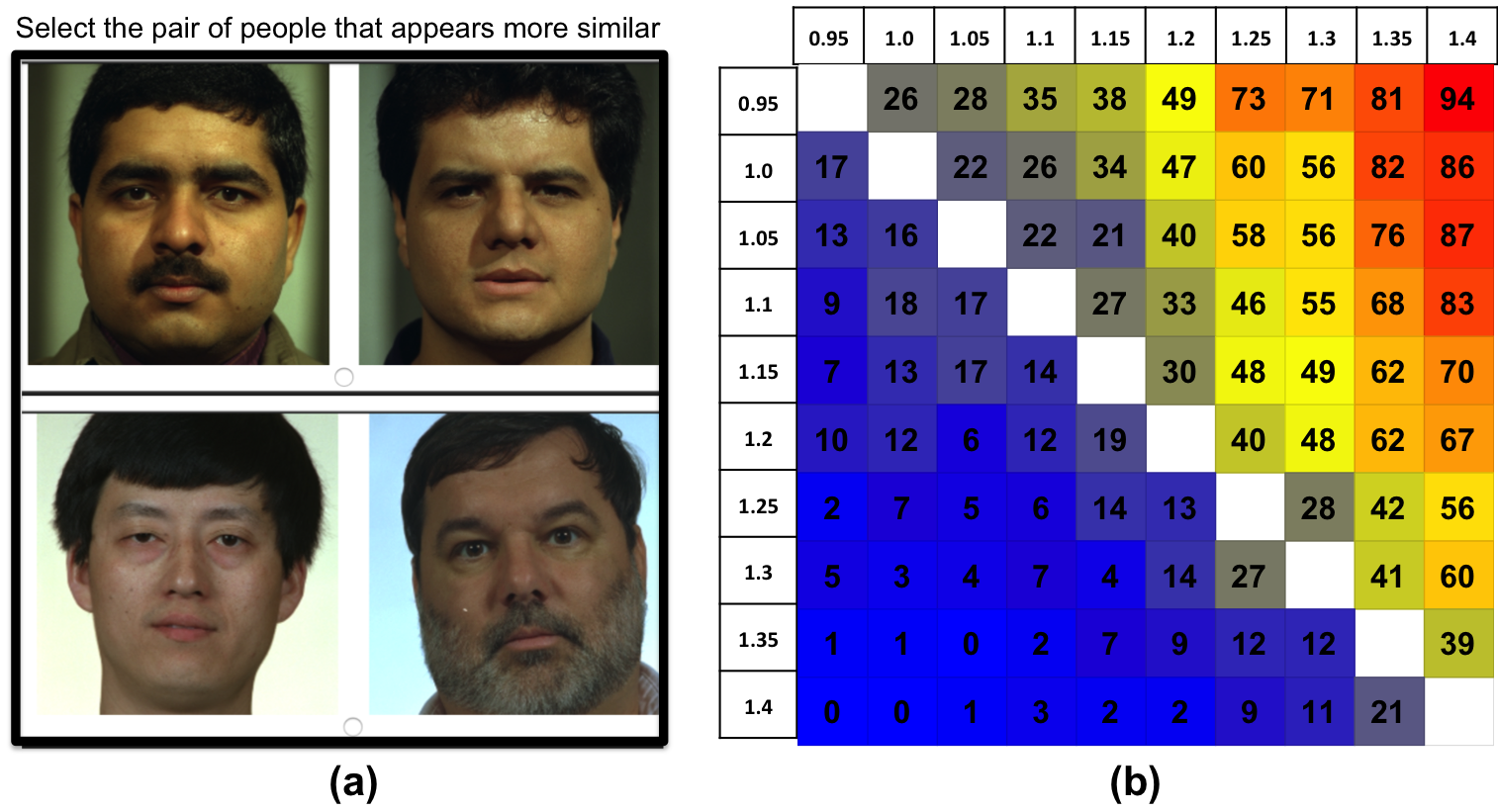}}

        \caption{Results showing the performance of the  VGG descriptor at the task of facial similarity. (a) An example of our Amazon Mechanical Turk task used for examining this relationship. (b) The number of times the row bin was chosen over the column bin (ignoring tests if there was not at least a 80\% agreement). The header signifies the distance upper bound for each bin.  More details can be found in Sec. \ref{sec:dataAnalysis}.}
        
        \label{fig:binMatrix}

        \vspace{-10pt}
\end{figure}

Using Amazon Mechanical Turk, we conduct an experiment where we present a pair of pairs to a worker and ask which pair  of faces appear more similar to each other. An example of the task is shown in Fig \ref{fig:binMatrix}. Each Mechanical Turk worker is presented with 10 random pairs of pairs, and each task is performed by 10 different workers.

\subsection{Data analysis}

\label{sec:dataAnalysis}

Results are presented in Fig. \ref{fig:binMatrix}(b), where each cell $a_{ij}$ at row $i$ column $j$ of the matrix is a comparison of a pair with distance $d_i$ to a pair of distance $d_j$. The bins are in increasing order (that is $i>j$ means that $d_i>d_j$). Fig. \ref{fig:binMatrix}(a) shows the number of tasks (out of 100) where pairs with distance $d_i$  were selected as more similar than pairs with distance $d_j$. We ignore comparisons where there was not at least an 80\% agreement among the workers (therefore $a_{ij}+a_{ji} \leq 100$ ). This means that at least 8/10 labelers agreed that a certain pair was more similar. We do this to ensure that we are examining meaningful pairs.

When first examining Fig. \ref{fig:binMatrix}(b) it is clear that there is a strong correlation between the distance in the recognition embedding space and the perceived similarity. For example, the fact that the top right corner has high values shows that when comparing a small-distance pair to a large-distance pair, the small-distance pair is almost always selected as the more similar. Even though these numbers do not reach 100, it is usually not because the larger bin pair was selected but simply because there was not 80\% agreement. This can be seen in the bottom left corner where most cells are approximately 0.

However, the top left corner of the matrix shows evidence that although similarity and recognition are similar, they may not necessarily be the same. In this region we are looking at tasks where both pairs have relatively small distances. Since the difference of the distances is small, many of the tasks do not yield an agreement of 80\% or above. However, when examining the pairs which do result in a high agreement we find that $a_{ij}$ is not much bigger than $a_{ji}$. That is, in many cases workers selected the pairs with higher distance as the more similar. In fact when adding the 5 smallest distance bins being compared to each other (The top left quarter of the matrix), and comparing the sum of the upper triangle (pairs of pairs which were labeled in accordance to their embedding distance) to the sum of the bottom triangle (those who were labeled inversely to what the embedding predicted), the accuracy is only 66.43\%.

Therefore, although \ref{fig:binMatrix}(b) clearly shows that there is a strong correlation between similarity and recognition, it appears to not work very well when looking at a group of images which distances are relatively small. That is, although the recognition embedding can clearly separate the ``not similar at all'' from the ``somewhat similar'', it does not do a great job at finding the most similar image in that group. This is an important deficiency since in many situations this would be the goal of face similarity. We would like our similarity measure to be able to select the ``most similar person'', just as face recognition is tasked with selecting one single identity.

\section{The Lookalike Network}

\label{sec:training}

In order to predict the perceived similarity of faces to more accurately reflect human opinion, we need to train the network specifically targeted for that task. However, it is much harder to collect a face similarity dataset than a face identity database. This is because one can extrapolate an identity of a face using different metadata such as captions or tags, as in \cite{guo2016ms}. This allows for large datasets with thousands of identities and millions of images. Since face similarity cannot be explicitly derived from such meta data it is infeasible to train a deep convolutional neural network for this task from scratch in an unsupervised or semi-supervised manner.

Therefore we decide to use a pre-trained network for face recognition and then fine-tune the weights from that initial state to perform well at the facial similarity task. This is a reasonable idea since we expect that many features which are useful for face recognition will also be useful for face similarity. In fact in Sec. \ref{sec:dataAnalysis} we have shown that the two tasks are highly correlated. Therefore we predict that starting with this pre-trained network and retraining it for our task should yield good results. We do this by adding a triplet loss layer on top of the original VGG-Face network. We then fine-tune this network on a new dataset that we collect that is targeted at capturing perceived facial similarity.

\subsection{Data Collection}

\label{subsec:datacollection}

As far as we know, there is no public database that includes information on facial similarity we decide to collect our own novel database. We design a task to collect the dataset with the following characteristics:

\begin{enumerate}
        
        \item The task should explicitly collect information regarding perceived facial similarity.
        
        \item The task should focus on images which are likely to be considered similar to one other. As discussed in Sec. \ref{sec:dataAnalysis} these are the images which the face recognition network's distance, when interpreted as a measure of facial similarity, gets wrong. In addition, we are much more interested in finding the most similar image amongst a group of somewhat similar images, than in finding examples of more similar faces to a query face from a pool a faces where none look particularly similar to the query at all. 
        
        \item Since perceived facial similarity is a subjective measure, we would like to have multiple workers complete the task.
        
        \item Since we are using triplet loss for training, we would like to be able to extract multiple triplets from a single task, thus maximizing the amount of data we have for training.
        
\end{enumerate}

\begin{figure}

        \centerline{\includegraphics[width=8.5cm]{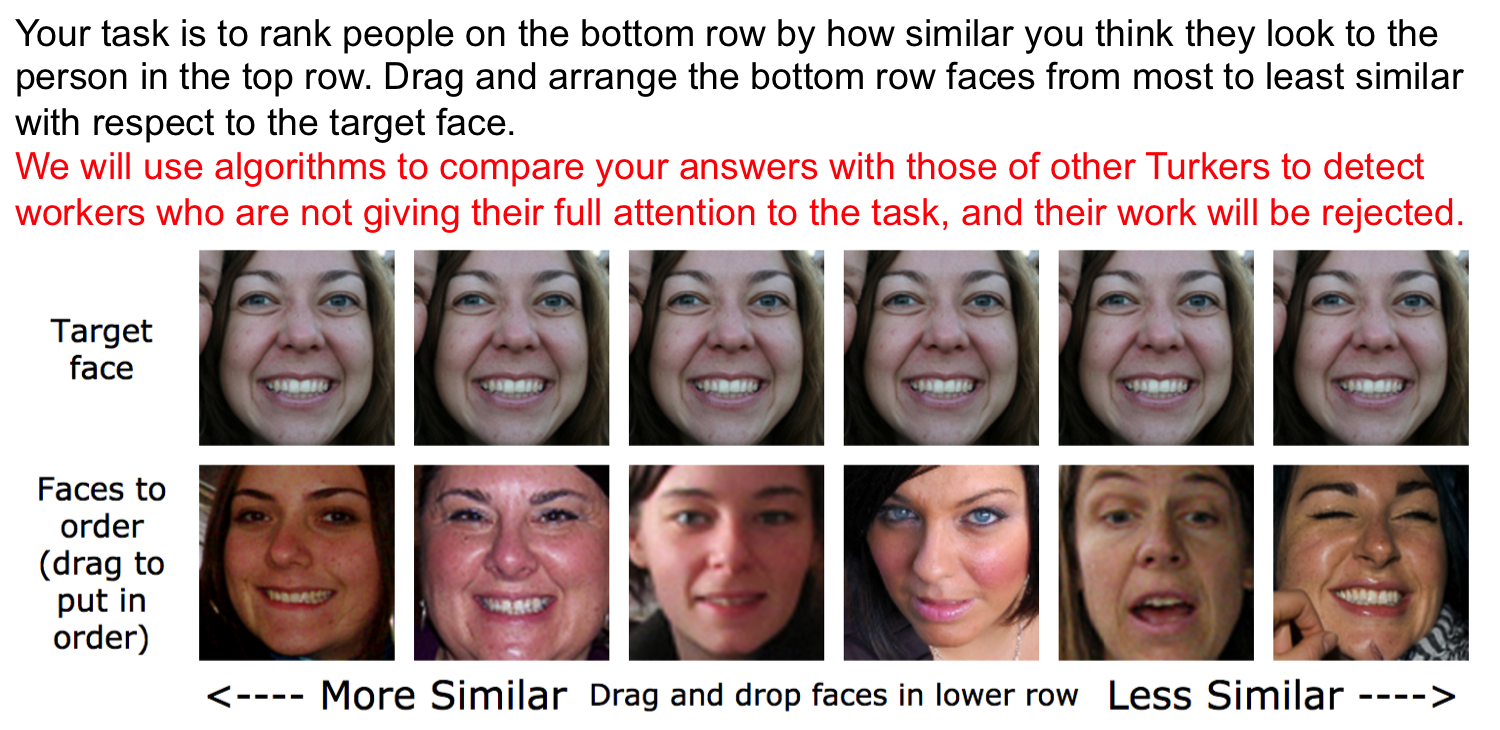}}

        \caption{An example of our Mechanical Turk Task. The worker is directed to drag/drop images from the bottom row such that images which are more similar to the top image are on the left, and images which are less are on the right.}
        
        \label{fig:taskexample}

        \vspace{-10pt}
\end{figure}

We therefore frame the task as a ranking task. Each task is composed of a reference query face ($I_0$), and the 6 most similar faces from a set of other identities based on the original VGG embedding distance ($I_{1 \ldots 6}$). We use this embedding distance since as shown in Sec. \ref{sec:dataAnalysis} these images are more likely to be considered as similar faces to the query face. We then present the images in the following manner. In the top row we present $I_0$ six times. Underneath we present $I_{1 \ldots 6}$ in a random order. The worker is then tasked with reordering images from the bottom row by dragging and dropping them so that the most similar image would appear on the left, while the most different image would appear on the right. The reason we present the query image multiple times is in order to make the face comparison easier to the worker (by just comparing images vertically). An example of the task can be seen in Fig. \ref{fig:taskexample}

We decided to use the Names100 dataset \cite{chen2013s} for this experiment for a few reasons. First, it is a larger dataset than the Color-Feret and thus gives us more images to work with. Second, as opposed to the Color-Feret dataset it is comprised of images ``in the wild'' and thus the network should generalize better to other datasets and images on the web. Third, as opposed to other datasets such as LFW \cite{LFWTech} and CelebA \cite{liu2015faceattributes} it is not comprised of celebrity photographs. We believe this is important since knowing a person might skew how similarity is judged. Finally, although the exact identity is not provided we can infer at least some identity from the names. Thus, we can ensure that images $I_0$ and $I_{1 \ldots 6}$ are not of the same identity by rejecting any image with the same name as $I_0$.

We randomly select 5000 images from the Names100 dataset. Then we use each of the 5000 images as a query image and select the 6 most similar images with different names based on the original VGG embedding, creating a total of 5000 Hits. We have 10 different workers complete each Hit. Since in a subjective task it is difficult to identify and remove lazy workers, we simply count how many images the worker rearranged and remove all results from workers who did not rearrange at least 1.5 images per hit (on average).

\subsection{Training}

Once we have collected the dataset, we retrain the network to specialize at the task of facial similarity, producing the Lookalike Network. We frame this as a ranking problem and seek to find an embedding which will better represent face similarity as compared with the face recognition embedding. We choose to use a triplet loss for the fine tuning of the network, where the loss is defined as:

\begin{equation}
L=\sum_{i=1}^{n}max(0,\lVert f(x_i^a)-f(x_i^p)\rVert - \lVert f(x_i^a)-f(x_i^n) \rVert + \alpha)
\end{equation}

Where $x_i=(x_i^a,x_i^p,x_i^n)$ represents a triplet of images, and $f(x) \in \mathbb{R}^d$ is the output of the network. The loss function attempts to learn an embedding in which the anchor images $x_i^a$ are closer to the positive images $x_i^p$ than the negative ones $x_i^n$ by at least $\alpha$. For image recognition \cite{wang2014learning} this is done by selecting positive datapoints from those with the same identity as the anchor, and negatives from those images from other class identities. In our case, we would like an image which workers ranked as more similar to the anchor face to be closer to the anchor than an image ranked more distant.

When performing training using triplet loss, the question of how to select triplets is an important one which can significantly effect the resulting model. We can select triplets from our Mechanical Turk task by using our query image as an anchor, and then using all pairs of similar images as positives or negatives. Since each task has six images, we have a total of ${{6}\choose {2}} = 15$ pairs. Each triplet has what we call a confidence level. The confidence level is defined as the percentage of people who ranked the positive image over the negative. Therefore, the positive and negative images are organized in such a way so that the confidence level is always above 0.5.

All these triplets may be considered to be hard triplets. Since we originally selected images for the tasks which had a low distance in the VGG embedding space, we know that they are all somewhat similar to the anchor (as discussed in Sec. \ref{sec:dataAnalysis}). Therefore, simply fine tuning on these examples might cause the network to make mistakes regarding simpler images. We therefore try an additional training setup which adds random easy triplets to the training data. The advantage is that we can do this for free without the need for manually annotating more data. Since we know that images with a large distance in VGG space will be almost surely be perceived as not similar (as shown in Sec. \ref{sec:dataAnalysis}), we can simply select one of those. In our implementation we simply select the positive as any one of the images from the Hit (small VGG distance) and as a negative we select a random image whose distance is larger than the median distance between the anchor and all other images in the dataset.

We use TensorFlow for training. We set $\alpha=0.05$ and a batch size to 32. We use stochastic gradient descent with an Adam Optimizer and a learning rate of $10^{-4}$. When training with additional random triplets, we set the probability of selecting a random triplet to 0.5.

\begin{figure*}

        \centerline{\includegraphics[width=16cm]{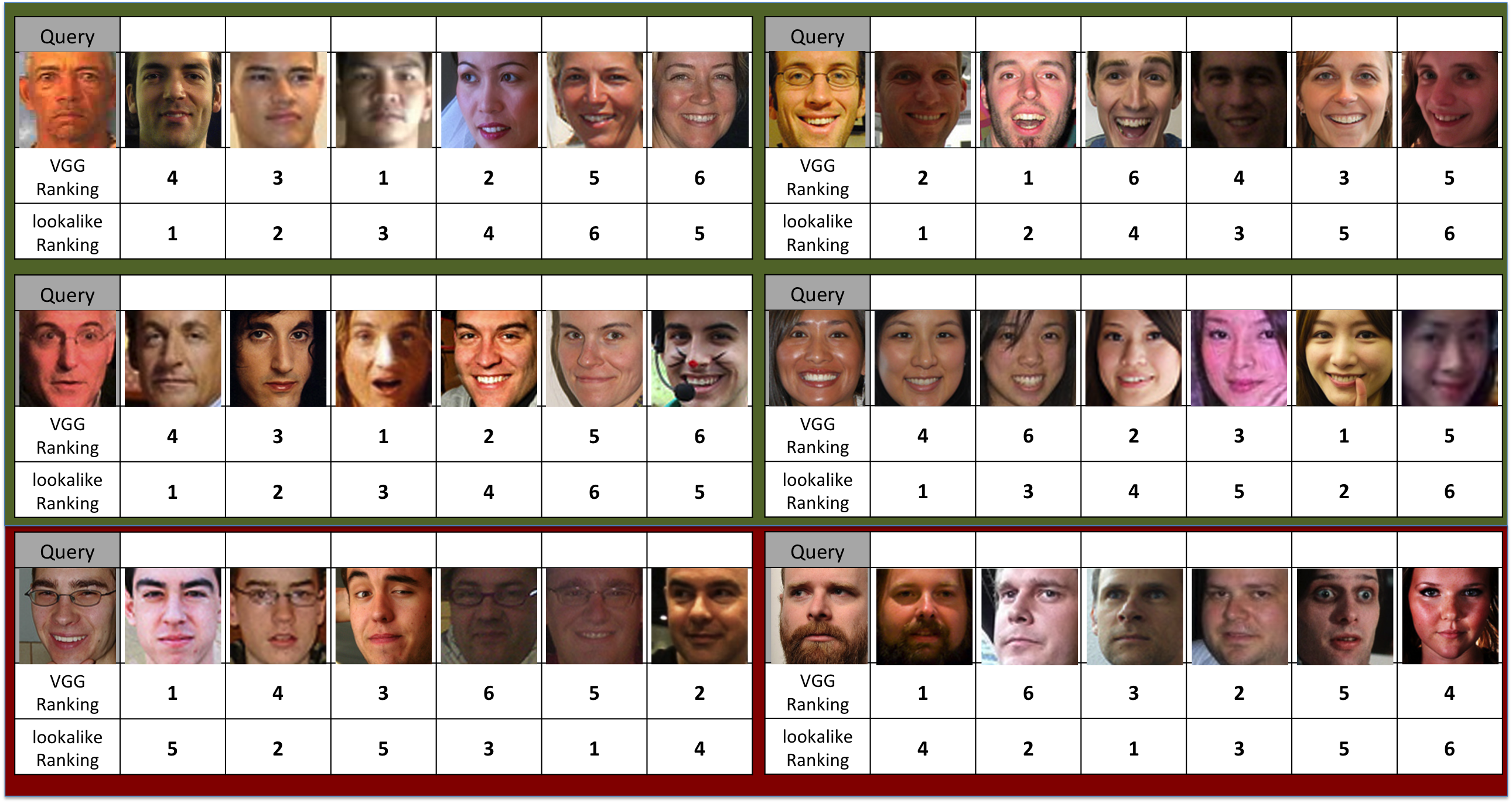}}

        \caption{Examples of the ranking results. The images are shown in the order of their ground truth ranking, i.e., the average ranked position by the Mechanical Turk workers). We then show the ranking by using the face recognition and our newly trained embedding distance. }
        
        \label{fig:results}

        \vspace{-5pt}
\end{figure*}

\section{Results}

\label{sec:results}

\begin{table}
       \footnotesize
        \renewcommand{\arraystretch}{1.5}        
        
        \begin{center}
                
                \begin{tabular}{|c||c|c|c|}
                        
                        \hline
                        
                        Training & \parbox[c][0.7cm][c]{1.9cm}{Testing on\\hard triplets} & \parbox[c][0.7cm][c]{1.9cm}{Testing on\\easy triplets} & \parbox[c][0.7cm][c]{0.8cm}{Total} \\ 
                        
                        \hline
                        
                        \hline
                        
                        \parbox[c][0.7cm][c]{1.5cm}{Original VGG Face} & 56.4\% & 100\% & 78.2\% \\
                        
                        \hline
                        
                        \parbox[c][0.7cm][c]{1.5cm}{Only hard triplets} & 66.7\% & 86.65\% & 76.68\% \\
                        
                        \hline
                        
                        \parbox[c][0.7cm][c]{1.5cm}{Include easy triplets} & 65.42\% & 97.15\% & 81.36\% \\
                        
                        \hline

                \end{tabular}

        \end{center}
                 \caption{Triplet accuracy. We compare the original face recognition network (second row) to our Lookalike network with two different training setups. One which solely uses triplets from the Hits (third row), while the other introduces random easy triplets as well (fourth row). We present the accuracy on both hard (second column) and easy (third column) triplets. Because we test on an equal number of hard and easy triplets, the last column is simply the mean of the first two columns. }
         \label{tab:resultsa}
        
        \vspace{-10pt}
\end{table}

We divide our dataset into a 90\%-10\% split to test our algorithm. We ensure that there are no identities overlapping between the two dataset splits. The results presented in this section are the averages taken over five splits using random sampling. First, in order to affirm that the Lookalike network is able to learn similarity we simply measure the triplet accuracy of the test set. Results are presented in Table \ref{tab:resultsa}. We show the results from the original VGG network, the Lookalike network trained strictly on  hard triplets, and the Lookalike network trained including easy triplets. In a similar fashion we measure the accuracy of each network on hard triplets and easy triplets.

As expected our network is able to do a much better job on hard triplets, increasing the accuracy by approximately 10\%. However, when training the network solely on them, the Lookalike network appears to get some easy triplets wrong. This is in contrast to the original VGG Face network which manages to easily reject random distant faces with 100\% accuracy. With training,  the network weights are updated without any penalty when originally distant faces are brought closer together. By adding easy triplets to the training set, this issue is remedied, and achieves a 97\% accuracy on easy triplets while still performing very well on hard ones.

Table \ref{tab:accuarcyat} presents our accuracy results for triplets in different ranges of confidence. It is preferable that if the worker agreement is higher (the more workers selected the positive over the negative), the accuracy of our algorithm should also be higher. One way to think of this is that if the Lookalike makes a mistake on a low worker agreement triplet, we displease fewer workers than if the network get a high worker agreement triplet wrong. The table shows that  the original VGG networks essentially does poorly in all different agreement levels (random is 50\%). Using the Lookalike network we do not only perform better in all different worker agreement levels, but our accuracy improves with the as the agreement gets higher which is the desired result. The fact that accuracy drops slightly for the 0.9-1 range could simply be because there are not many triplets in that range.

In order to show that our algorithm performs better not just on individual triplets, but at ranking as a whole, we examine a few ranking metrics on the original images selected by the VGG network vs. our reordering. Table \ref{tab:topranked} shows the probability of the top ranked face being in the top $k$ positions. As shown, the Lookalike network generally places the top ranked image in a higher position. In addition we also calculate the Normalized discounted cumulative gain (NDCG) by using the equation:

\begin{equation}
NDCG_6= \frac{1}{ODCG_6}\sum_{i=1}^{6} \frac{2^{rel_i}-1}{log_2(i+1)} 
\end{equation}

Where $rel_i$ is defined as 6 minus the average position the image was placed (therefore relevance scores range from 6-0) and $ODCG_6$ is the optimal discounted cumulative gain (when the images are ordered by their average position). The Lookalike network produces an increases the NDCG from 0.844 to 0.891.

We present visual examples of our results in Fig. \ref{fig:results}. Although our algorithm does not always get the exact order correct, there are a few patterns which seem to hint at why our algorithm does better. For example, in the top row examples the original VGG network seems to rank people from the opposite sex as similar. This may not hurt in the recognition task because if the image is already far enough in embedding space the gender is irrelevant. However, for similarity we have found that mostly people do not consider faces from opposite genders as similar (although it does occasionally occur). This is likely one of the cues that Lookalike is able to pick up. Although the gender issue is one factor, it is certainly not enough to justify the entire improvement. In fact since the images are already somewhat similar there are not many sets which include opposite genders. The second row shows examples in which gender does not play a role since all images are of the same gender. In fact there is not clearly an attribute we can point to which makes the images more or less similar. These examples emphasize the fact that our network is able to learn something more subtle than simply counting attributes that match (see further discussion in Sec. \ref{sec:discussion})

Finally, the last row shows some failure cases. In the left example the Lookalike network seems to rank people with glasses higher than the others since the query image has glasses as well. From our observation, it is usually true that workers tend to consider people with glasses more similar to each other. However, this example clearly shows that this is not always true. On the right we appear to have the opposite problem in which our network did not rank the bearded person highly, even though the query is bearded and the VGG network correctly ranked it as the top image.

In order to ensure that Lookalike generalizes to other datasets we conduct two additional experiments. First, we return to the data we collected in Sec. \ref{subsec:datacollection}. Our goal is to examine if the Lookalike network performs better on pairs image pairs that are nearly the same distances apart in VGG face recognition embedding space. We examine the images from 5 bins with the smallest distance (corresponding to the top left corner of Fig. \ref{fig:binMatrix}(b)) and measure the accuracy of the original embedding vs. the Lookalike embedding to the worker consensus. The original VGG embedding achieves 66.43\% match to the worker consensus, while Lookalike achieves an increase to an accuracy of 78.1\%. Thus, we are able to fix some of the concerns raised in Sec. \ref{sec:dataAnalysis}.

\begin{figure}

        \centerline{\includegraphics[width=8cm]{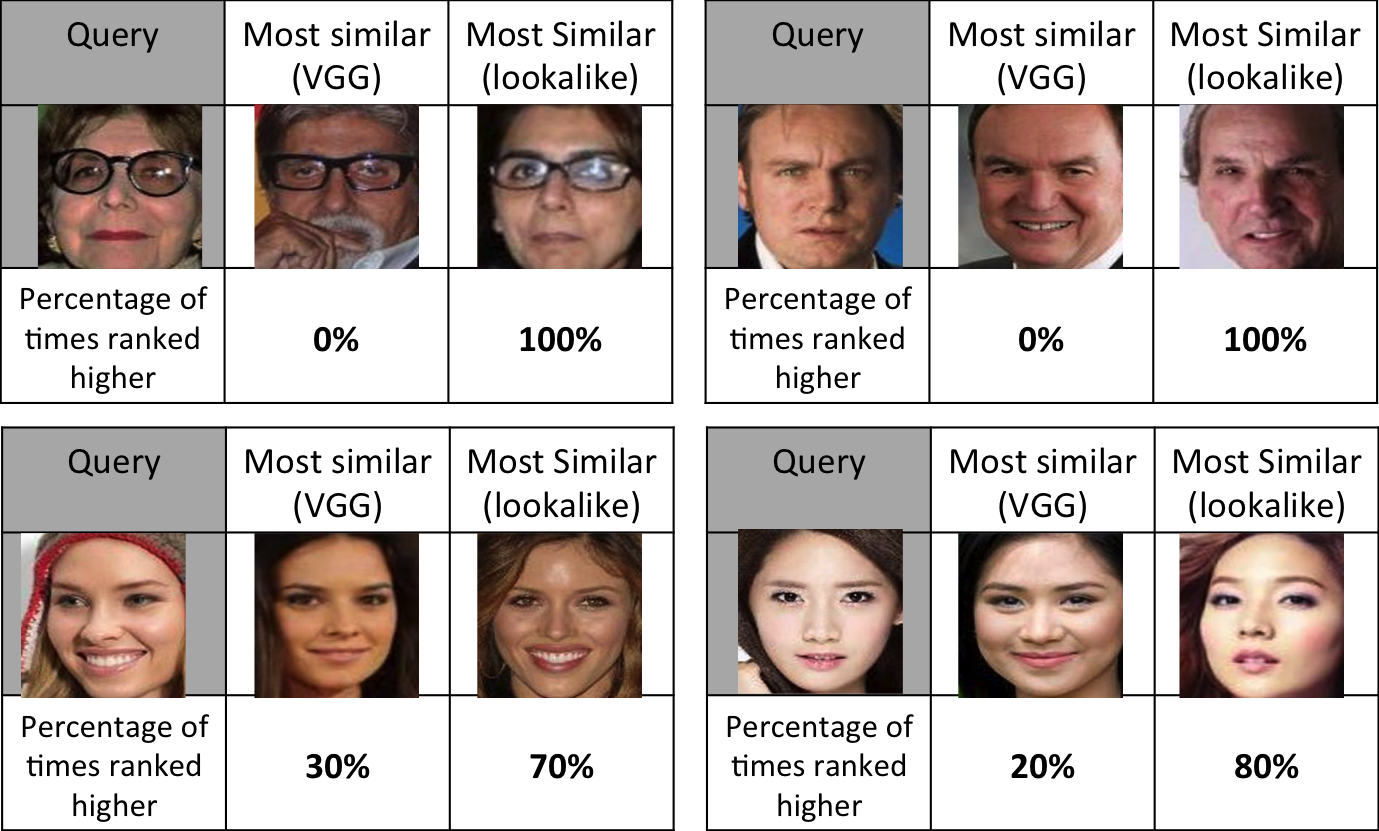}}

        \caption{Examples from our CelebA experiment. We show four examples of a  query image and the most similar image as judged by the VGG network and our Lookalike network. Also shown is the percent of workers which judged the image to be more similar to the query. }
        
        \label{fig:resultsceleba}

        \vspace{-10pt}   
\end{figure}

We run an additional experiment on the CelebA dataset. As in our training case we select the 6 most similar images using the VGG network and then reorder them using the Lookalike network. We then select the top image from each list and have workers arrange the photos in order of similarity. We then examine the average position of each image for each Hit. In this dataset we see the same effect as in the Names100 dataset. That is, the top image selected by the Lookalike network is placed higher than the VGG embedding selection 58.33\% of the time, while the  opposite (the VGG selected face being closer to the query than Lookalike's selection) is only true 39.33\% of the time. (The other 2\% are ties). A few examples are shown in Fig.\ref{fig:resultsceleba}.

\begin{table}
        
        \footnotesize

        \renewcommand{\arraystretch}{1.5}        
        
        \begin{center}

                \begin{tabular}{ |c||p{0.9cm}|p{0.9cm}|p{0.9cm}|p{0.9cm}|p{0.9cm}|}
                        
                        \hline
                        
                        Network & 0.5-0.6 & 0.6-0.7 & 0.7-0.8 & 0.8-0.9 & 0.9-1 \\
                        
                        \hline
                        
                        \hline
                        
                        \parbox[c][0.7cm][c]{1.5cm}{Original VGG Face} & 52.54\% & 53.48\% & 55.39\% & 54.77\% & 54.88\% \\
                        
                        \hline
                        
                        \parbox[c][0.7cm][c]{1.5cm}{Lookalike Network} & 58.90\% & 64.09\% & 68.92\% & 73.99\% & 73.5\% \\                
                        
                        \hline
                        
                \end{tabular}

        \end{center}
        
        \caption{Accuracy on triplets in a specific confidence range. The accuracy of the Lookalike network seems to be closely correlated with the agreement/disagreement among subjects. }
        
        \label{tab:accuarcyat}

\vspace{-10pt}        
\end{table}

\begin{table}
        
        \footnotesize

        \renewcommand{\arraystretch}{1.5}        
        
        \begin{center}

                \begin{tabular}{ |r||p{0.8cm}|p{0.8cm}|p{0.8cm}|p{0.8cm}|p{0.8cm}|}
                        
                        \hline
                        
                        $k=$ & 1 & 2 & 3 & 4 & 5 \\
                        
                        \hline
                        
                        \hline
                        
                        \parbox[c][0.7cm][c]{1.4cm}{Original VGG Face} & 21.6\% & 38.4\% & 57.6\% & 76.2\% & 88.4\% \\
                        
                        \hline
                        
                        \parbox[c][0.7cm][c]{1.4cm}{Lookalike Network} & 33.2\% & 50.77\% & 66.86\% & 81.92\% & 92.44\% \\                
                        
                        \hline
                        
                \end{tabular}

        \end{center}
        
        \caption{Precision of the top-ranked image being in the top $k$ images for $k=1 \ldots 6 $ as ranked by the two networks.}
        
        \label{tab:topranked}

\vspace{-10pt}        
\end{table}

\section{Discussion}

\label{sec:discussion}

There can be several reasons why the Lookalike network ends up outperforming the original reporposed VGG face embedding at the task of understanding perceived facial similarity. First, it could be that the task of face recognition and face similarity are the same, and our fine tuning is simply improving both tasks by adding more training data. We therefore recalculate the accuracy of the Lookalike network on LFW \cite{LFWTech}. Since all images in the Names100 dataset \cite{chen2013s} we use for training are cropped in a certain manner, we crop the LFW images in the same way to ensure that our method is not performing poorly because of dataset bias (for this experiment we only use LFW images with landmarks provided).

The original VGG network achieves a 99.7\% AUC while the lookalike network only reaches 97.69\%. This shows that our network has learned a new task that actually slightly impaired its utility at the task of face verification, rather than simply improving face recognition.

Another interesting claim would be that our network is simply learning facial attributes and is just looking for faces which share the most amount of attributes in common. This claim would be more difficult to analyze since there is no comprehensive and agreed upon list of nameable facial attributes, and the importance of each attribute with respect to the concept of facial similarity is unclear. However, we decided to do a first order approximation of this by looking at the attribute distribution of the most similar faces as determined by the VGG embedding and our Lookalike network.

We use the CelebA dataset \cite{liu2015faceattributes} since it contains ground truth attributes for each image. We then randomly select one image per identity and use both networks to select the 6 most similar people in the database. In addition, we also select the six most similar people by attribute. That is, we use a binary attribute vector (of length 40) to describe each image, and then look for the six images with the smallest Hamming distance. We then examine the average Hamming distance in attribute space between the top six images and the query face. The VGG network has an average distance of 0.185 while the Lookalike has an average distance of 0.179, i.e., slightly more similar in terms of matching attribute count. For comparison, the six closest images in attribute space are only an average of 0.05 away. Therefore, it appears that the Lookalike network does not select faces with significantly more attributes in common, and therefore is probably learning something deeper than just attributes.  

\section{Conclusion}

\label{sec:future}

In this work, we introduce the novel task of learning perceived facial similarity. We show through data collection that facial recognition and perceived facial similarity are related, yet distinct tasks with the former being based on categories (identity) and the latter based on relative similarities. We describe a method to training a deep neural network to perform this specific task, including data collection and triplet selection methods. Finally we present our results which show that our algorithm, the Lookalike network, outperforms the face recognition baseline at the task of predicting which faces will appear more similar to a human. We also show that this improvement generalizes across multiple face datasets. We believe that this provides strong evidence that face similarity is an important topic and distinct from face recognition. We hope that the dataset that we have collected to train Lookalike, which we will share, will inspire further work in this area.

This work raises many questions which we believe are important  to investigate. First, there are more tasks to address in the realm of perceived facial similarity. For example, although in this work we attempt to find a more similar instance of a face, it would be interesting to aggregate distances across multiple photos or videos of people.

In addition, since there is not always a consensus on which two images are more similar, a deeper investigation can be done to determine advantageous ways to integrate this level of agreement into the triplet loss training. That is, should the margin depend on how confident we are that the positive is closer than the negative? Should the loss be higher if we violate a more confident triplet? These questions will not only improve the task of image similarity, but can translate to other domains in which these issues exist such as relative attributes, image memorability, and more.

%\subsection*{Acknowledgements}
%We thank Wassim Gharbi and Thanh Vu of Lafayette College for helping with data collection and initial experiments.

{\small
\bibliographystyle{ieee}
\bibliography{refs}
}

\end{document}